# *A State-of-the-Art of Semantic Change Computation*

Xuri Tang

Huazhong University of Science and Technology

E-mail: xrtang@hust.edu.cn

## Abstract

This paper reviews state-of-the-art of one emerging field in computational linguistics — semantic change computation, proposing a framework that summarizes the literature by identifying and expounding five essential components in the field: diachronic corpus, diachronic word sense characterization, change modelling, evaluation data and data visualization. Despite the potential of the field, the review shows that current studies are mainly focused on testifying hypotheses proposed in theoretical linguistics and that several core issues remain to be solved: the need for diachronic corpora of languages other than English, the need for comprehensive evaluation data for evaluation, the comparison and construction of approaches to diachronic word sense characterization and change modelling, and further exploration of data visualization techniques for hypothesis justification.

## 1 Introduction

In the field of Computational Linguistics (CL), the past decade has witnessed an emerging research topic — semantic change computation, exemplified by research works such as Sagi, Kaufmann, and Clark (2009), Michel, Yuan, Aiden, Veres, Gray, Team, Pickett, Hoiberg, Dan, and Norvig (2011), Rohrdantz, Hautli, Mayer, Butt, Keim, and Plank (2011), Jatowt and Duh (2014), Mitra, Mitra, Maity, Riedl, Biemann, Goyal, and Mukherjee (2015), Frermann and Lapata (2016), Hamilton, Leskovec, and Dan (2016), Tang, Qu, and Chen (2016), Dubossarsky, Grossman, and Weinshall (2017) and others. The surge of interest in the topic is attributed to development inside CL and out of it. Within CL, the availability of large scale diachronic language data and the technology development of Word Sense Induction (D. Lin, 1998; Nasiruddin, 2013; Navigli, 2009) have provided access to information of diachronic semantic change of words in automatic means, thus paving the way to empirical investigation of regularities and mechanisms in semantic change. Outside CL, the development of Internet technologies, the web-based social media in particular, has accelerated the change of language (Crystal, 2006). New words and new usage of words are popping up in the course of web-based communications.

The emergence of the research topic is not surprising, given the fact that semantic change has long been of interest to both academic circle and the general public. Based on the observations that language is always changing (Hale, 2007: 3) and language is a dynamic flux (Beckner, Ellis, Blythe, Holland, Bybee, Ke, Christiansen, Larsen-Freeman, Croft, and Schoenemann, 2009; Mantia, Licata, and Perconti, 2017; Massip-Bonet, 2013), linguists had put forward different theories and models in searching for rules and regularities in semantic change, such as the Diachronic Prototype Semantics (Geeraerts, 1997, 1999), the Invited Inference Theory of Semantic Change (Traugott and Dasher, 2002), and semantic change based on metaphor and metonymy (Heine, Claudi, and Hünnemeyer, 1991; Sweetser, 1990a, 1990b). The knowledge of semantic change has also been interesting to the public because people need to be aware of such change in their daily language use.

A large number of websites and books are devoted to etymology studies (Jatowt et al., 2014).

Although at the moment no published research is available on how the incorporation of semantic change might impact the performance of NLP systems, the potential of semantic change computation to Natural Language Processing (NLP) is great (Jatowt et al., 2014; Mitra et al., 2015). Researchers have realized that "understanding how words change their meaning over time is key to models of language and cultural evolution" (Hamilton et al., 2016). Semantic change computation deals with the dynamic change of semantics, including the phenomenon of polysemy, a perennial challenge for NLP. A better understanding of polysemy should help improve any NLP task that involves semantic processing. For instance, applications for semantic search can increase the relevance of the query result by taking into account the new senses of words (Mitra et al., 2015; Yao, Sun, Ding, Rao, and Xiong, 2017). In addition, the knowledge of trends and tendencies discussed in semantic change computation, including the knowledge obtained in Culturomics (Michel et al., 2011), is also essential in NLP tasks. Semantic change computation also plays a role in social computing, such as determining the diachronic change of the popularity of brands and persons and the most famous athletes at different times (Yao et al., 2017).

This paper intends to review the literature of semantic change computation by proposing a framework that identifies core components in the task: diachronic corpora, diachronic word sense characterization, change modelling, evaluation and data visualization. State-of-the-art of each component is expounded in the paper, together with relevant insights and findings from theoretical linguistics and cognitive linguistics. It is our hope that a systematic review that integrates both theoretical and computational legacy could benefit research in the future.

## 2 Framework of Semantic Change Computation

Theoretical studies of semantic change address three sub-topics: semasiology, onomasiology and path of semantic change (Geeraerts, 1997; Traugott et al., 2002: 25-26). In semasiology, the linguistic form is kept constant, and the focus is on the development of polysemy. Onamasiology, instead, is focused on the development of linguistic representation of particular semantic domains such as COLOR and INTELLECT. The third sub-topic concerns paths of change across conceptual structures, as evidenced by particular semasiological changes.

Current literature of semantic change computation is mainly devoted to semasiology, because linguistic forms can be conveniently identified in corpora to enable a thorough investigation of the phenomena. Onomasiology poses a challenge to semantic change computation because of the difficulty in formally representing a semantic domain and in obtaining comprehensive information of semantic domains in large scale unannotated corpora. As for studies of the paths of semantic change, they are based on onomasiology and semasiology. As such, the framework presented in Figure (1) concerns mainly semasiology. It summarizes research paths that most studies in the literature take.

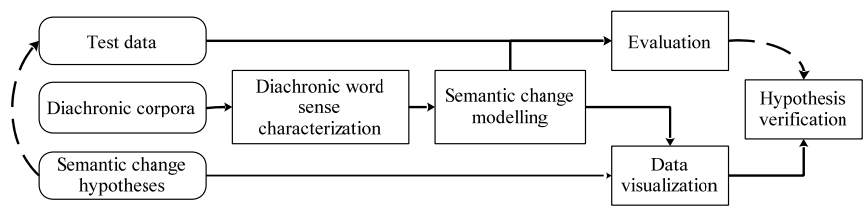

Figure 1. Framework of Semantic Change Computation

Figure (1) shows that the field is still in its early phase, focused on verifying general hypotheses proposed in theoretical studies of semantic change. The beginning component in the framework is *Semantic change hypotheses*. Several general regularities and rules of semantic change are addressed in the literature, such as the detection of novel senses (Erk, 2006; Frermann et al., 2016; Lau, Cook, McCarthy, Newman, and Baldwin, 2012), categorization of semantic change (Mitra et al., 2015; Tang et al., 2016), correlation between frequency and semantic change (Dubossarsky et al., 2017; Hamilton et al., 2016), and semantic change and polysemy (Hamilton et al., 2016).

The framework presents two paths to *Hypothesis verification*, indicated by the arrows in the figure. Beginning with *Semantic change hypotheses*, one path proceeds directly to *Hypothesis verification* with information obtained from *Semantic change modelling*, using visualization tools (Martin and Gries, 2009; Rohrdantz et al., 2011) and mathematical reasoning (Dubossarsky et al., 2017; Hamilton et al., 2016). The other path transforms hypothesis verification into categorization tasks. It first exemplifies hypotheses by constructing evaluation data that contain semantic change categories such as sense shift, sense narrowing and sense broadening and then evaluating the change modelling methods against the evaluation data. The evaluation results are finally used to verify the hypotheses. Most studies in the literature follow this approach, although the scales of data and categories of semantic change under investigation vary in the studies.

Both paths in the framework rely on two essential components in semantic change computation: *Diachronic word sense characterization* (DWSC) and *Semantic change modelling* (SCM). The component of DWSC gathers diachronic semantic information of words from corpora. It generally involves two steps: (1) inducing word senses from corpora with unsupervised methods such as clustering; (2) summarizing usage of words and transforming word sense representations so that sensible comparison can be computationally conducted among different time intervals.

The component of SCM uses computational models to interpret hypotheses of semantic change with information gathered from DWSC. Metaphysical understanding of the concept *change* determines the type of computational model used in the component. Section 5 in this review distinguishes two types of metaphysical understanding of *change*: Cambridge Change and McTaggart Change and expounds their corresponding computational models. The intrinsic nature of the hypothesis under investigation should also be considered in choosing computational models. For instance, semantic change categorization can be modelled with machine learning tools such as Support Vector Machine (Tang et al., 2016) and correlation between semantic change and frequency can be modelled with linear regression (Hamilton et al., 2016).

The rest part of the paper reviews state-of-the-art of the five essential components in the framework, namely *Diachronic corpora*, *Diachronic word sense characterization*, *Semantic change modelling*, *Evaluation* and *Data visualization*.

## 3 Diachronic Corpora

Diachronic corpora provide empirical resources for semantic change computation. The construction and choice of diachronic corpora involve consideration of factors such as size, balance, and representativeness of the corpora in particular. As argued in Sinclair (2005), only those components of corpora that have been designed to be independently contrastive should be contrasted in a study. For semantic change computation, only those corpora that are designed to be informative of semantic change should be used, and only those semantic aspects that are contrastive in the corpora should be studied.

Table (1) lists most of the corpora used in the literature of semantic change computation. The most frequently used corpus is the Google Books Ngram Corpus (Y. Lin, Michel, Aiden, Orwant, Brockman, and Petrov, 2012), due to its size, range of time, types of language included in the corpus and public availability. The corpus has been available at http://books.google.com/ngrams since 2010. Its 2012 Edition is obtained from more than 8 million books, or over 6% of all books ever published in history. The corpus consists of ngrams (n ≤ 5) of the following format:

Table 1. Diachronic corpora in Semantic Change Computation

| Names | Language and Times | Used in |
| --- | --- | --- |
| Google Books Ngram Corpus | English, French, Spanish, German, Chinese, Russian and Hebrew (From 1500s to 2008) | Michel et al. (2011), Gulordava and Baroni (2011), Hamilton et al. (2016), Wijaya and Yeniterzi (2011), Jatowt et al. (2014), Yang and Kemp (2015), Dubossarsky et al. (2017) |
| Corpus of Historical American English (COHA) | American English (1810s-2000s) | Neuman, Hames, and Cohen (2017), Hamilton et al. (2016) |
| Google Books Syntactic Ngram Corpus | English (1520-2008) | Mitra et al. (2015) Goldberg and Orwant (2013) |
| Helsinki Corpus of English Texts | Old English (till 1150) Middle English (1150-1500) Early Modern English (1500-1710) | Sagi et al. (2009), Sagi, Kaufmann, and Clark (2011), |
| TIME corpus from BYU | American English (1923-2006) | Martin et al. (2009) |
| New York Times Corpus | American English (1987-2007) | Rohrdantz et al. (2011) |
| Corpus from New York Times | American English (1990-2016) | Yao et al. (2017) |
| DATE corpus[1] | English (1700-2100) | Frermann et al. (2016) |
| Newspaper Corpus from People's Daily | Modern Chinese (1946-2004) | Tang et al. (2016) |
| Newspaper Corpus from Le Monde | Modern French (1997-2007) | Boussidan and Ploux (2011) |
| Parole Corpus for modern Swedish and Swedish Literature Bank for Swedish in the 19th century | | Cavallin (2012) |
| British National Corpus for late 20th century and ukWaC for 2007 | | Lau et al. (2012) |
| Twitter Corpus, 1% of the Twitter data from 2012 to 2013 | | Mitra et al. (2015) |

---

[1] The corpus is compiled by collocating documents from (1) the COHA, (2) the training data provided by the DTE task organizers (http://alt.qcri.org/semeval2015/task7/index.php?id=data-and-tools) in SEMEVAL 2017 and (3) the portion of the CLMET3.0 CORPUS.

ngram TAB year TAB match_count TAB volume_count NEWLINE

The Google Books Syntactic Ngram Corpus is based on the English portion of the ngram corpus and are POS-annotated and dependency parsed.

However, we should be aware of its limitations when we use Google Books Ngram Corpus. With the corpus, the biggest window size for observation in WSI is $\pm 4$, so it is not suitable for semantic change occurring in long distance dependency. In addition, language variation factors such as genre and social events can neither be directly studied with the corpus, which is unfortunate because social factors play significant roles in semantic change (Hollman, 2009; Labov, 1994). The corpus is mainly representative of formal language because it is built from books, not spoken language. Corpora like COHA, Helsinki Corpus of English Texts, and those corpora compiled from newspapers are better candidates for studies of social factors in semantic change.

Table (1) also shows that the most studied language is English, including British English and American English. Other languages studied in the literature are Chinese, French, and Swedish. Although the statistical techniques might be universally applicable to all languages, the patterns of language change might differ for different languages because language change is motivated by pragmatics (Traugott et al., 2002), which is part of the culture.

## 4 Diachronic Word Sense Characterization

Diachronic corpora are all not sense annotated. Therefore, Diachronic word sense characterization (DWSC) is introduced to automatically capture comparable semantic state-of-affairs of the target word (named global semantics in this paper) at different time intervals by collecting information from slices of diachronic corpora. It can be performed in one or two steps. Two-step DWSC involves

(1) specifying the instance sense, namely the sense of the target word occurring in a sentence and (2) computing the global semantics of the target word including types of sense and the prevalence of the senses in a slice of diachronic corpus. One-step DWSC is performed by directly obtaining the global semantics of one time interval or all the time intervals within one model. For instance, the SCAN model proposed in Frermann et al. (2016), which is a Bayesian model of sense change based on topic modelling, obtains simultaneously comparable global semantics of the words for all time intervals.

This component is closely related to Word Sense Induction (WSI). Many techniques based on distributional semantics (Bullinaria and Levy, 2012; Firth, 1957; Harris, 1954; Levy, Goldberg, and Dagan, 2015; Weaver, 1955) are also adopted in semantic change computation to induce word senses in diachronic corpora. Word contexts provide important information not only for WSI, but also for detecting semantic change (Sagi et al., 2009). Therefore, the global semantics of a target word is often represented by word vectors, which could be high dimensional sparse vectors or low dimensional vectors.

However, DWSC differs from WSI in that the global semantics obtained from different time intervals should be comparable. This paper has opted for categorizing approaches of DWSC into two categories: approaches based on explicit word vectors that are directly comparable and word-embedding based approaches that require alignment to make word vectors comparable. They are detailed in the following two sections.

### 4.1 Approaches Based on Explicit Word Vectors

The approaches falling into this group use explicit word vectors to characterize diachronic word senses. In an explicit word vector, each cell in the vector corresponds to association strength between the target word and a context word, namely the word occurring in the context of the target word. The context word can be identified in different ways. It can be the collocates forming collocation with the target word. For instance, Tang et al. (2016) represents the instance sense as a tuple $c = (w_t, w_m)$, in which $w_t$ is the target word and $w_m$ is the co-occurring noun word with the strongest association strength with $w_t$ within a window size of $\pm 9$. In Cavallin (2012), the context word is a noun or a verb that form verb-object relation with $w_t$. Mitra et al. (2015) used bigram relationship to specifying the context word.

Within this group, various methods are proposed to obtain global semantics of the target word. In Tang et al. (2016), the global semantics of $w_t$ is summarized as the Average Mutual Information (Entropy) of all the possible senses in the corpus, as denoted below:

$$\zeta^{w_t} = \begin{bmatrix} C \\ P(C) \end{bmatrix} = \begin{bmatrix} c_1, c_2, \dots, c_i, \dots c_n \\ p(c_1), p(c_2), \dots, p(c_i), \dots p(c_n) \end{bmatrix} = H(C) = -\sum_{i=1}^{n} p(c_i) \log p(c_i) \quad (1)$$

where $c_i$ represents one sense of $w_t$, $p(c_i)$ denotes its probability. The global semantics in the form of Entropy is an indicator of the usage scope of the word and is therefore directly comparable. Higher entropy indicates a wider scope of usage. In Cavallin (2012), the global semantics is a list of words that are collected from instance senses and ranked according to their association strength with the target word. The ranking information encoded in this form makes it convenient for manual analyses of lexical semantic change. In Mitra et al. (2015), the global semantics is obtained by integrating the instance senses into a thesaurus-based graph (Figure (2)), and is induced with graph-clustering techniques.

Positive Point Mutual Information (Bullinaria and Levy, 2007; Bullinaria et al., 2012) is also used in the literature to obtain global semantics. With this approach, the global semantics of the target word is obtained by computing the Point Mutual Information (PMI) $I(w_c; w_t) = \log(p(w_c|w_t)/p(w_c))$ between the target word $w_t$ and words in its context $w_c$, and by keeping those $w_c$ with $I(w_c; w_t) > 0$. According to Bullinaria et al. (2012), keeping those with $I(w_c; w_t) > 0$ helps keep the words that co-occur more frequently with the target word and get rid of those that co-occur with low frequency.

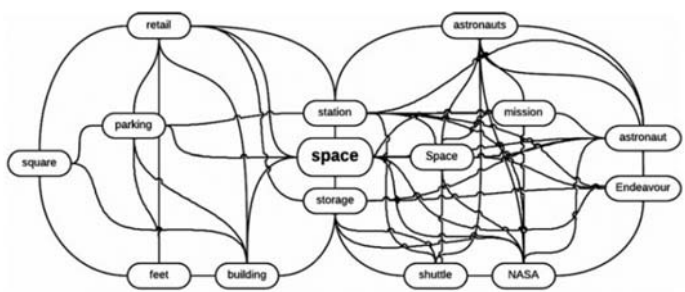

Figure 2. Word co-occurrence network for the word *space* (Mitra et al., 2015)

The major advantage of using explicit word vectors for global semantics is that these vectors are conveniently interpretable and directly comparable across different time intervals. Nevertheless, because the dimensions of these vectors are often very high, the problem of data sparseness may lead to poorer performance compared with Word-Embedding approaches.

### 4.2 Approaches Based on Word-Embedding

Word-embedding based approaches map high-dimension word vectors to low-dimension vectors to obtain global semantics. According to the co-occurrence scope considered in the operation, the techniques can be further divided into two sub-types: those relying on topic modelling and those relying on local word context.

The approaches relying on topic modelling use co-occurrences of words within documents to characterize the semantics of the words, such as Latent Semantic Indexing (LSA) (Deerwester, Dumais, Furnas, Landauer, and Harshman, 1990) and Latent Dirichlet Allocation (LDA) (Blei, Ng, and Jordan, 2003). Take LDA as an example. The idea of LDA is that documents are represented as random mixture over latent topics, where each topic is characterized by a distribution over words (Blei et al., 2003). The idea could be summarized in the following equation:

$$p(w|d) = p(w|t) * p(t|d) \qquad (2)$$

With LDA, words are represented as their distributions over topics. Lau et al. (2012) compared LDA (that needs to specify the number of topics) and Hierarchical Dirichlet Process (HDP) (Teh, Jordan, Beal, and Blei, 2006) (that needs not to specify the number of topics) in their ability in obtaining global semantics of words. Other studies that use topic modelling include Rohrdantz et al. (2011), Jatowt et al. (2014) and others.

Topic modelling can be extended to incorporate time to model the evolution of topics over time, represented by models such as the Topic-Over-Time model (TOT) (Wang and Mccallum, 2006) and dynamic topic models (Blei and Lafferty, 2006). TOT is used in Wijaya et al. (2011) for DWSC. TOT is a time-dependent extension of LDA that parameterizes a continuous distribution over time associated with each topic. By thus doing, the topics also take on information of time.

The SCAN model proposed by Frermann et al. (2016) is a dynamic Bayesian topic model, described in Figure (3). For each target word, a SCAN model is created that represents the meaning of the word with two distributions: a K-dimensional multinomial distribution over word senses $\phi$ and a V-dimensional distribution over the vocabulary $\psi^k$ for each word sense. As a generative model, SCAN starts with a precision parameter $\kappa^\phi$, a Gamma distribution with parameters *a* and *b* that controls the extent of meaning change of $\phi$. A three time step model involves the generation of three sense distributions $\phi^{t-1}$, $\phi^t$, and $\phi^{t+1}$, at times *t-1*, *t* and *t+1*. The sense distribution at each time generates a sense *z*, which in turn generates the context words for the target word. Simultaneously, for each time of *t-1*, *t* and *t+1*, the word-sense distributions $\psi^{k,t-1}$, $\psi^{k,t}$ and $\psi^{k,t+1}$ are generated, each of which also generates the context words for the target word. The prior is an intrinsic Gaussian Markov Random Field which encourages smooth change of parameters at neighboring times. The inference is conducted with blocked Gibbs sampling.

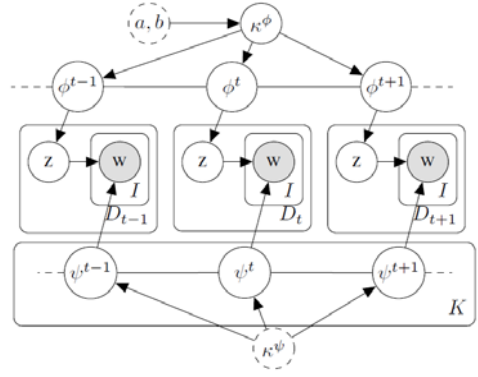

Figure 3. Three time step SCAN model (Frermann et al., 2016)

Word embedding techniques that rely on local context of the target words include Word2vec (Mikolov, Sutskever, Chen, Corrado, and Dean, 2013), generalized SVD and Glove (Pennington, Socher, and Manning, 2014) etc. The scope of context is restrained within a window size. For instance, the objective of the Skip-gram model is to maximize the average log probability of the following:

$$\frac{1}{T}\sum_{t=1}^{T}\sum_{-c\leq j\leq c, j\neq 0}\log p(w_{t+j}|w_t) \tag{3}$$

where c is the window size of the context and $w_t$ is the target word (Mikolov et al., 2013). The obtained word vector is characterized by words within the context of size *c*. Hamilton et al. (2016) compared the performance of PPMI, generalized SVD and Skip-gram model on two tasks: detecting known semantic shifts and discovering shifts from data. The research concludes that PPMI is worse than the other methods, SVD performs best on detecting semantic shifts, and Skip-gram performs best on discovering semantic shifts. Dubossarsky et al. (2017) uses PPMI and general SVD to evaluate laws of semantic change.

The word vectors obtained with embedding techniques such as general SVD, Word2Vec, and Glove cannot be used directly for DWSC. Different from explicit word vectors obtained by PPMI, these techniques map the original stochastic distributions of the words to lower dimensions, which will preclude comparison of the same word across time (Hamilton et al., 2016). Therefore, word vectors of different time intervals obtained with these techniques have to be aligned to become comparable. Various approaches are proposed for the purpose. Based on the assumption that word vector spaces are equivalent under a linear transformation and that the meanings of most words did not shift over time, Kulkarni, Alrfou, Perozzi, and Skiena (2014) adopts a linear transformation $W_{t'\mapsto t}(w) \in \mathbb{R}^{d\times d}$ that maps a word vector $\phi_t$ to $\phi_{t'}$ by solving the following optimization problem:

$$W_{t'\mapsto t}(w) = \underset{W}{\operatorname{argmin}} \sum_{w_i\in k-NN(\phi_t,(w))}\left\|\phi_{t'}(w_i)W - \phi_t(w_i)\right\|_2^2 \tag{4}$$

where $k - NN(\phi_t, (w))$ defines a set of k nearest words in the embedding space $\phi_t$ to the word $w$. Hamilton et al. (2016) proposes to use orthogonal Procrustes to align the word vectors obtained via general SVD and Skip-Gram word2vec. By Procrustes analysis, the word vectors are translated, scaled and rotated for alignment. Given $W^{(t)} \in \mathbb{R}^{d \times |V|}$ as the matrix of word embeddings learned at year *t*, the research align the matrices by optimizing:

$$R^{(t)} = \arg\min_{Q^T Q = I} \left\| QW^{(t)} - W^{(t-1)} \right\|_F \tag{5}$$

where $R^{(t)} \in \mathbb{R}^{d \times d}$ and $||\cdot||_F$ is the Frobenius Norm. The solution can be obtained using an application of SVD.

Similar to TOT model, Yao et al. (2017) proposed a diachronic word embedding model that integrates the word embedding learning process and the alignment operation by solving the following joint optimization problem:

$$\min_{U(1),\ldots,U(T)} \frac{1}{2}\sum_{t=1}^{T}\left\|Y(t) - U(t)U(t)^T\right\|_F^2 + \frac{\lambda}{2}\sum_{t=1}^{T}\left\|U(t)\right\|_F^2 + \frac{\tau}{2}\sum_{t=2}^{T}\left\|U(t-1) - U(t)\right\|_F^2 \tag{6}$$

where $Y(t)$ is the word embedding matrix, $U(t)$ is the desired temporal word embedding for time *t*, and $\lambda, \tau > 0$. The model enforces alignment across all time slices and helps avoid propagation of alignment errors.

There are also word-embedding based studies that make use of both topic modelling and word-context information, such as the semantic density model proposed in Sagi et al. (2009). The study first obtained a vocabulary of word vectors by applying Latent Semantic Indexing (Deerwester et al., 1990) on a term-term matrix of size 40,000x2000 (40,000 words as vocabulary, 2000 words as content words) filled with weighted co-occurrence counts of words, resulting in a reduced 40,000x100 matrix, with each item in the vocabulary associated with a 100-dimensional vector. Given a target word $w_t$, the instance sense of $w_t$ is the context vector which is the normalized sum of the vectors associated with the words occurring with $w_t$ in the instance sentence within the window size $\pm15$. The global semantics of $w_t$ is denoted by the density or the average pairwise similarity of the instance senses collected from the given corpus. The similarity of two instance senses is measured by cosine similarity.

There is currently no comprehensive comparison of all these approaches to diachronic word sense characterization, partly due to the fact that there is no large scale evaluation data available for the purpose. Therefore, the nature of these approaches requires further research.

## 5 Semantic Change Modelling

DWSC gathers the global semantics of the target word at each time interval. How the global semantics of different time intervals is used to interpret semantic change and testify hypotheses of semantic change is the task of the component *Semantic change modelling*. Semantic change is a type of *change*. The prerequisite of semantic change computation is to clarify how *change* is metaphysically defined. Different metaphysical understandings of *change* cover different phenomena of semantic change and require different techniques in semantic change modelling.

### 5.1 Metaphysical Understandings of Semantic Change

The literature has reported two types of metaphysical understanding of change: Cambridge Change (Geach, 1969: 71-72; 1979; Mortensen, 2016), and McTaggart Change (Broad, 1938: 297). The definition of Cambridge Change is given below:

> *An object O is said to 'change' . . . if and only if there are two propositions about*
>
> *O, differing only in that one mentions an earlier and the other a later time, and one*

*is true and the other false. (Geach, 1979: 91)*

Many ordinary thoughts of change, such as change of color from 'red' to 'blue', and change of weather from 'hot' to 'cold' are of this type. It implies an intrinsic difference at two points along the dimension of time. The metaphysical understanding of McTaggart Change originates from John McTaggart, which is defined below:

> *"The thing changes from $q_1$ to $q_2$" is completely analyzable into a statement of the kind "There is a certain series of successive events so interrelated that it counts as the history of a certain thing [X], $e_1$ and $e_2$ are two successive adjoined phases in this series and $e_1$ has Q in the form $q_1$, while $e_2$ has Q in the form $q_2$ . . ." (Broad, 1938: 297)*

In McTaggart's view, a change occurring at two points of time does not constitute as a change. Instead, it establishes itself as a change when two successive events have different qualities. Both Cambridge Change and McTaggart Change agree that different qualities lead to change. They differ in how persistent the quality difference lasts. Cambridge Change is justified so long as the difference is detected at two distinctive points of time, while McTaggart Change maintains that change is identified should the difference exist in the successive course of time, not just two slices of time.

  Whether to adopt Cambridge Change or McTaggart Change in studies of semantic change relates to an important concept in language — convention. A conventionalized regularity is a regularity known and accepted at communal level and observed in almost any instance where the regularity applies (Lewis, 1969: 42). Because Cambridge Change considers only two points in the course of time, the semantic change based on such metaphysical understanding does not consider the course of time and therefore does not take convention into serious consideration. McTaggart Change, however, is based on events happening in the course of time and therefore incorporates conventionalization in its study.

Correspondingly in lexicography, convention serves to distinguish two types of semantic change: institutionalization and topicalicality (Fischer, 1998: 16). Institutionalization refers to the process by which a semantic change is integrated into language norm and becomes a convention, while topicality refers to the phenomenon that a word is used in connection with current affairs for a short period. For instance, Landau (2001) discussed the example *chat group* and speculated that the phrase may disappear in a couple of years with development of technology. Therefore, the phrase *chat group* is more of topicalicality than of institutionalization. From this perspective, Cambridge Change and McTaggart Change cover different semantic change phenomena. Cambridge Change covers both topicalicality and institutionalization, while McTaggart Change has a narrower scope and mainly concerns the institutionalization process that involves conventionalization.

Studies in theoretical semantic change generally adopt McTaggart Change as the metaphysical understanding of semantic change (Andersen, 1989; Fortson, 2003; Traugott et al., 2002). In the literature of semantic change computation, both Cambridge Change and McTaggart Change are adopted.

## 5.2 Computation Models based on Cambridge Change

Semantic change computation based on Cambridge Change mainly relies on similarity comparison. Given $\zeta_{t_1}$ and $\zeta_{t_2}$ as the global semantics of the target word *w* at time $t_1$ and $t_2$, the semantic change of *w* is indicated by the similarity measure as below:

$$\theta = \text{Similar}(\zeta_{t1}, \zeta_{t2}) \tag{7}$$

Various similarity measures can be used for the purpose, depending on how the global semantics is represented. For instance, Gulordava et al. (2011) use cosine similarity (Equation (8)) to measure two word vectors obtained from two decades to decide whether a word has undergone a semantic change. If $\zeta_{t1}$ and $\zeta_{t2}$ carry information about the probability of senses, comparison can be used to capture the novelty of senses using Formula (9), as proposed in Lau et al. (2012)

$$\cos(\zeta_{t1}, \zeta_{t2}) = \frac{\zeta_{t1} \cdot \zeta_{t2}}{|\zeta_{t1}||\zeta_{t2}|} \tag{8}$$

$$\text{Nov}(w) = \frac{\zeta_{t2} - \zeta_{t1}}{\zeta_{t1}} \tag{9}$$

Cambridge Change can also be interpreted as comparison of distances to a center. The idea is based on the proposal in Geeraerts (1999), who hypothesized that changes in the meaning of a lexical item are likely to be changes with respect to the prototypical "center" of the category. Dubossarsky, Tsvetkov, Dyer, and Grossman (2015) propose to use the centroid of the global semantics of the words belonging to the category as the "center" of the category. Based on word2vec and Google Books Ngram Corpus, the research obtained the distance of the words to the centroid of the category as illustrated in Table (2). A semantic change is identified when the distance of a word to the center of the category widens.

Table 2. Illustration of centroid-based distance

| shutters | windows | doors | curtains | blinds | gates |
|----------|---------|-------|----------|--------|-------|
| 0.04     | 0.05    | 0.08  | 0.1      | 0.11   | 0.13  |

Cambridge Change can be used to detect novel senses. Researchers have proposed various ways to compute sense novelty on the basis of the ratio of sense distribution, as shown in Table (3).

Table 3. Various ways to compute sense novelty

| | |
|---|---|
| Lau et al. (2012) | $\text{Novelty}_{\text{Ratio}}(s) = \dfrac{P_f(s)}{P_r(s)}$ <br> $P_f(s)$ and $P_r(s)$ are the proportion of usages of a given word corresponding to sense s in the focus corpus and reference corpus |
| Cook, Lau, McCarthy, and Baldwin (2014) | $\text{Novelty}_{\text{Diff}}(s) = p_f(s) - p_r(s)$ <br> $P_f(s)$ and $P_r(s)$ are the proportion of usages of a given word corresponding to sense s in the focus corpus and reference corpus |
| Cook et al. (2014), (Frermann et al., 2016) | $\text{Novelty}_{\text{Topic}}(s) = \sum_{w \in W} p(w\|s,t)$ <br> $p(w\|s,t)$ is the conditional probability of w given the induced sense s at time t and W is a collection of keywords identifying a certain topic. |

The major concern facing Cambridge-Change based models is conventionality. Comparison of global semantics of two time intervals can answer the question whether a semantic change occurs or not, but it cannot tell whether the change has been conventionalized. Mitra et al. (2015) shows that the comparison method flagged 2498 candidates with new semantic senses, but only 537 of them had been observed to be stable.

## 5.3 Computation Models based on McTaggart Change

The models based on McTaggart Change rely on time series for semantic change investigation. Most studies in the literature fall into this group, as seen in Frermann et al. (2016), Yang et al. (2015), Hilpert and Perek (2015), Jatowt et al. (2014), Wijaya et al. (2011), Tang et al. (2016) and many others. There are generally two ways to construct the time series. One way is to use the explicit global semantics of words obtained from corpora to form a time series, as below:

$$\xi_{1,n}^{w} = [\zeta_{t_1}, \zeta_{t_2}, \zeta_{t_3}, \ldots, \zeta_{t_n}] \qquad (10)$$

where $\zeta_{t_i}$ denotes the global semantics of the word *w* at time *t_i*, observed over some time intervals from time $t_1$ to time $t_n$. The time interval could be a century, a decade, a year, a month, or a day. For instance, the global semantics obtained in Tang et al. (2016) is the Average Mutual Information between the target word and its collocates. The other way is to construct a time series from the differences between the global semantics of the starting time (or ending time) and those of other times, denoted as below:

$$\xi_{1,n}^{w} = [\Delta(\zeta_{t_2}, \zeta_{t_1}), \Delta(\zeta_{t_3}, \zeta_{t_1}), \dots, \Delta(\zeta_{t_n}, \zeta_{t_1})\} \tag{11}$$

where $\Delta(\zeta_{t_i}, \zeta_{t_1})$ denotes the difference between $\zeta_{t_i}$ and $\zeta_{t_1}$. The time series constructed this way is mainly used for the analyses of change pattern. Cosine similarity is often used to obtain the difference between word vectors, as in Kulkarni et al. (2014).

With McTaggart Change, it is possible to apply analyzing techniques of time series to semantic change computation. For instance, Sample Autocorrelation Function (sample ACF) (Brockwell and Davis, 2002: 18) can be used to assess whether there is a trend of change in the time series and whether the trend consists of different stages. For illustration, given the time series data for the Chinese word *tou$^{51}$ming$^{35}$* from Tang et al. (2016) in Figure (4)[2], the sample ACF in Figure (5) shows that there are two sequences of decaying, indicating that there is a trend of change consisting of two stages. Note that data with Sample ACF below $\pm 1.96/\sqrt{n}$ (n is the number of data) is considered to contain no trend, since 1.96 is the .975 quantile of the standard normal distribution (Brockwell et al., 2002: 20).

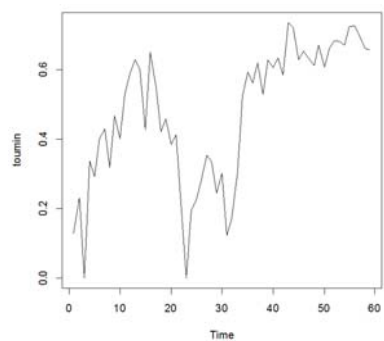

---

[2] Both this figure and the following one are extracted from Tang et al. (2016)

Figure 4. Time Series Data for the Chinese word *tou[51]ming[35]*

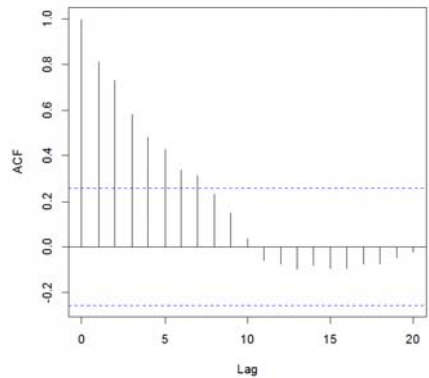

Figure 5. Sample ACF with $0 \leq \text{lag} \leq 20$ of the time series in Figure (4) with boundary $\pm 1.96/\sqrt{59}$

Compared with Cambridge-Change based models, the models based on McTaggart Change incorporate conventionality in semantic change computation, which enables more elaborated studies of semantic change, such as semantic change categorization.

The statistic nature of global semantics obtained in DWSC determines the type of categorization it can be used for. As discussed in Section 4, global semantics comes in different forms. Those obtained with LSA (as in Jatowt et al. (2014)), PPMI (as in Hamilton et al. (2016) and Dubossarsky et al. (2017)), or Skip-Gram model (as in Yao et al. (2017) and Hamilton et al. (2016)) do not distinguish between senses of the word. Although the difference obtained by contrasting the global semantics of different time intervals can be used for semantic change detection, it does tell whether the usage scope has increased or decreased. Therefore, global semantics of this nature cannot be used for more granulated categorization such as sense broadening and sense narrowing.

If the global semantics of a word contains information of different senses of the word, such as the entropy used in Tang et al. (2016), or word vectors obtained with LDA (Lau et al., 2012; Rohrdantz et al., 2011), TOT (Wijaya et al., 2011), and SCAN (Frermann et al., 2016), machine learning techniques can be applied to detect semantic change categories such as sense broadening, sense narrowing, pejoration, amelioration, even metaphorical change and metonymic change. For instance, Tang et al. (2016) makes use of curve fitting and Support Vector Machine to categorize semantic changes.

It is odd that the model of semantic change is not much discussed, although many studies have adopted the McTaggart-Change based approaches. One model, namely the S-Shaped Curve, is detailed in Tang et al. (2016). The S-Shaped Curve is defined by the logistic function below:

$$\zeta = \frac{e^{k+st}}{1+e^{k+st}} \qquad (12)$$

where $\zeta$ is the global semantics of the target word. It varies according to the time variable $t$. The two constants $k$ and $s$ are to be estimated with curve fitting techniques. This model has been confirmed in studies of language change in general (Bailey, 1973; Prévost, 2003; Zuraw, 2006), in grammar change (Kroch, 1989), in sound change (Labov, 1994), and in Culturomics (Michel et al., 2011). The two constants $k$ and $s$ in Formula (12) are closely associated with the intrinsic properties of the particular semantic change that the target word is undergoing. Tang et al. (2016) has used these constants to investigate semantic change categories such as metaphorical change, metonymical change, coining of new words, sense broadening and sense narrowing.

In addition, conventionality is also an issue not much discussed in the literature, although the McTaggart-Change based models enable such investigation. A new usage gains its status as a novel sense only when it is conventionalized as well as being frequently used. Tang et al. (2016) discusses the issue very briefly by proposing to compute the conventionality of a sense with the following:

$$\text{Conventionality}_s^T = AveragePrevalence(s,T) \times DiaSpan(s,T) \times DiaDensity(s,T) \quad (13)$$

The formula considers factors such as the average frequency of the sense *s* during the ***T*** period (*AveragePrevalence*), the duration of *s* in *T* (*DiaSpan*), and the frequency of *s* occurring in different time intervals(*DiaDensity*) in ***T***.

In sum, the McTaggart-Change based approaches to semantic change use time series to model semantic change. Although more information is available with the approach, there is still a need to explore how semantic change should be modelled and how computational models relate to those hypotheses proposed in theoretical linguistics.

## 6 Evaluation and Evaluation Data Construction

For the component of *Evaluation* in the framework, the general measures such as precision, recall and F-score can be adopted to measure the performance of semantic change computation models. The main challenge in the component is the construction of evaluation data. Although the past decade has witnessed the trend from case studies to the construction of evaluation data in the literature, several issues remain to be solved, including the categories of semantic change and principles of data collection.

The studies of semantic change serve different purposes. To explore the principles of semantic change, or to construct linguistic knowledge resources, it might require in-depth analyses of semantic change phenomena, including changes occurring to individual senses, conventionality of the senses, prevalence of senses, and even motivations of semantic change. On the other hand, Natural Language Processing applications are mainly concerned with detection of novel senses and dynamic update of lexical knowledge. Therefore, different types of evaluation data are needed to meet different interests.

## 6.1 Categories of semantic change

In theoretical linguistics, there are different categorization schemes of semantic change (such as Geeraerts (1983), Blank and Koch (1999), Bloomfield (1933) and many others). The categorization scheme proposed in Bloomfield (1933) is one of the most popular schemes. We propose to formalize the scheme proposed by Bloomfield in terms of connotation change, as illustrated in Figure (6). In five of the six categories in the figure, the original sense of the word is retained. But the sense can be widened (thus the category of widening) or narrowed (thus the category of narrowing). New senses can be added to the word. The added sense can be of the same semantic domain (thus the category of metonymy) or of different domains (thus the category of metaphor). The category of semantic shift specifies the cases in which a new sense is substituted for its original one. Along with the sense substitution, the emotional coloring of the word might be changed from commendatory to derogatory (thus the category of pejoration) or from derogatory to commendatory (thus the category of amelioration).

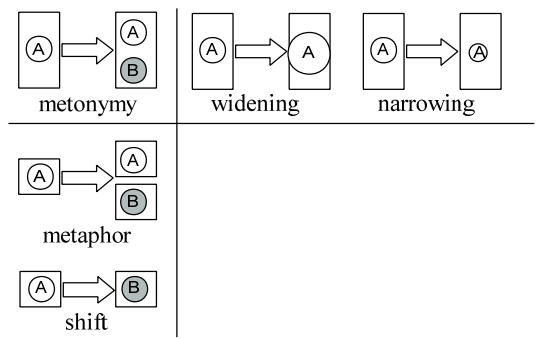

Figure 6. Schematic diagram of semantic change categories for theoretical exploration

The phenomena of widening and narrowing are discussed in most studies, such as Sagi et al. (2009) and Gulordava et al. (2011). Novel senses, namely the birth of lexical senses are discussed in details in Erk (2006), Lau et al. (2012), Mitra et al. (2015), Frermann et al. (2016), Tang et al. (2016) and others. Pejoration and amelioration are dealt with in Jatowt et al. (2014). Using words from SentiWordNet[3] as constitutes in context, the research conducts case studies of *aggressive*, *propaganda*, *fatal* in Google Books Ngram Corpus and shows that *aggressive* has undergone amelioration and that *fatal* and *propaganda* have undergone pejoration. Semantic shift is discussed in Wijaya et al. (2011), with examples such as *gay* and *awful*. Detection of semantic change motivated by metonymy and metaphor is discussed in Tang et al. (2016).

The studies of semantic change can also add the ability of time awareness to NLP systems and help improve performance in disambiguating word senses. Along this track, one possible categorization scheme is proposed in Mitra et al. (2015). The research proposed four types of semantic change based on change in semantic clusters: *split*, *join*, *birth* and *death*. Because splitting of one clustering into two and joining of two clusters are more of the results of clustering algorithm than of semantic change, we propose to modify the category schema into three categories: *non-change*, *birth* and *death*, as illustrated in Figure (7).

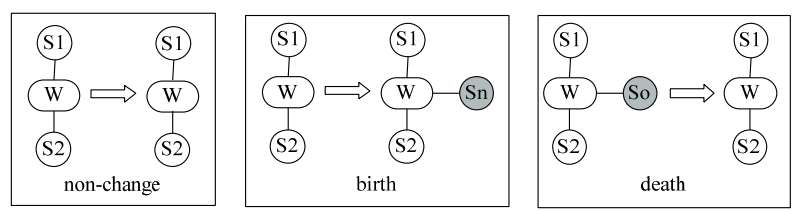

---

[3] http://sentiwordnet.isti.cnr.it

Figure 7. Schematic diagram of semantic change categories for NLP

## 6.2 Methods of Evaluation Data Collection

Large scale evaluation of meaning change is notoriously difficult (Frermann et al., 2016). This explains the reason why many studies in semantic change computation have opted for case studies, and why there exists no standard evaluation data in the field. However, researchers have made efforts to collect and construct evaluation data in their studies. Two methods are identified in the literature: rating based data collection and dictionary based data collection.

### *6.2.1 Rating Based Data Collection*

Gulordava et al. (2011) reported constructing an evaluation data of 100 words from the American English portion of Google Books Ngram. The 100 words are selected from 10,000 randomly picked mid-frequency words in the corpus and are rated by humans. During the rating process, human raters were asked to rank the resulting list according to their intuitions about the change of the word in the last 40 years on a 4-point scale (0: no change; 1: almost no change; 2: somewhat change; 3: changed significantly). Although the average of pair-wise Pearson correlations of the 5 participants was .51, the reliance on rating could be problematic due to the age and personal experience of the participants.

### *6.2.2 Dictionary Based Data Collection*

Dictionaries, particularly those dictionaries that have published different editions over history, might be the most reliable source for evaluation data collection. For instance, the Oxford English Dictionary has three editions: the first edition beginning in 1928, with supplementary volumes added in later years; the second edition in 1989; the third Edition is on the way[4]. Dictionaries are reliable because their compilation process is often prudent, during which more than one lexicographers should participate in gathering evidence to support their decisions as to whether to include or exclude word senses.

Several studies in the literature collect evaluation data on the basis of dictionary. Rohrdantz et al. (2011) used 2007 Collins Dictionary for later senses of words and the English WordNet (Fellbaum, 1998) and 1987 Longman Dictionary for earlier senses of words. The research conducted case studies of the words *browse*, *surf*, *messenger*, *bug* and *bookmark*, finding meaning extension of these words. Mitra et al. (2015) used dictionaries for both manual and automatic evaluation. For manual evaluation, they used New Oxford American Dictionary as the gold and manually checked 69 words against the dictionary, obtaining a precision of 60% for the birth category and a precision of 57% for the split/join category. For automatic evaluation, the research used WordNet 3.0 for reference. A sense mapping is constructed to map senses of clusters to senses of synsets in WordNet. Since they model semantic change on the basis of Cambridge Change, data of 1090-1953 are compared with data of other periods such as 1954-1972, 1973-1986, and 1987-1995. The range of precision is between 30% and 65% for birth, split and join. In constructing a dataset of diachronic sense differences to date, Cook et al. (2014) made use of two editions of Concise Oxford Dictionary: the 1995 Thompson Edition and the 2005 Soanes and Thompson Edition. They first identified a number of lemma by selecting words whose entries contain words such as *code*, *computer*, *internet*,

---

[4] Information obtained via the Oxford English Dictionary webpage: http://public.oed.com/history-of-the-oed/oed-editions/.

*network*, *online*, *program*, *web* and *website*. Tang et al. (2016) used 14 dictionaries to compile evaluation data, among which there are ten new word dictionaries and four general dictionaries.

## 7 Data Visualization

Semantic change computation often involves data that are large, complex and multi-dimensional, as it deals with words that are often in large numbers along the dimension of time. Researchers have opted for data visualization for better illustration, description, and exposition of semantic change regularities, as in Martin et al. (2009), Hilpert et al. (2015), Michel et al. (2011), Rohrdantz et al. (2011) and others.

A common visualization method is to plot different senses over time, as illustrated in Figure (8). The senses are represented by terms obtained with topic modelling. Figure (8b) differs from Figure (8a) in that Figure (8b) also shows the percentage of the senses (therefore the prevalence of the senses) of the word at each time interval. Plots with sense distribution over time make it convenient to observe semantic changes such as broadening, narrowing, semantic shift, pejoration and amelioration. In Figure (8b), the topic of "*air*, *joy*, *love*, *heart* …" is gradually detached from the word *transport*, indicating people's attitude towards *transport* changes from pleasant to neutral, which is a process of pejoration.

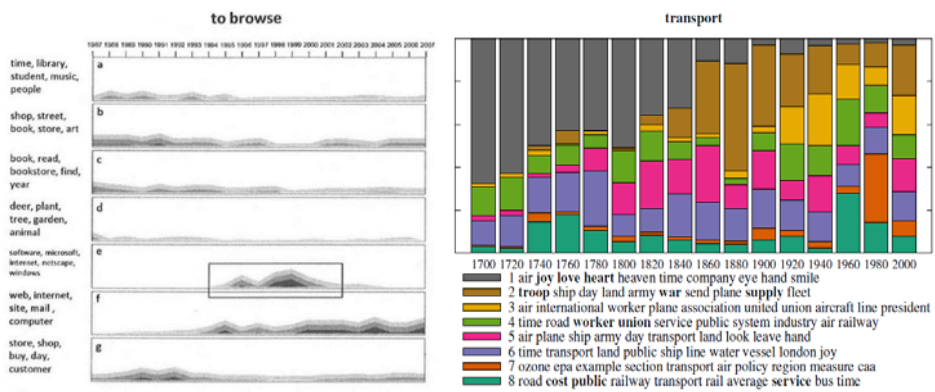

(a)        (b)

Figure 8. Sense distribution over time. Figure (a) is retrieved from Rohrdantz et al. (2011), Figure (b) from Frermann et al. (2016).

Hilpert et al. (2015) introduces a visualization method that animates scatterplots over time to explain the semantic change of a construction "many a NOUN", as depicted in Figure (9). The spots in the scatterplot stand for nouns found in the construction. They are plotted according to their semantic distance. The sizes of the spots indicate the frequency of the nouns found in the construction and the colors indicate different semantic categories. With the animated scatterplots, it is easy to see that the construction gradually acquires popularity by collocating with more words, particularly those words of the semantic domain TIME.

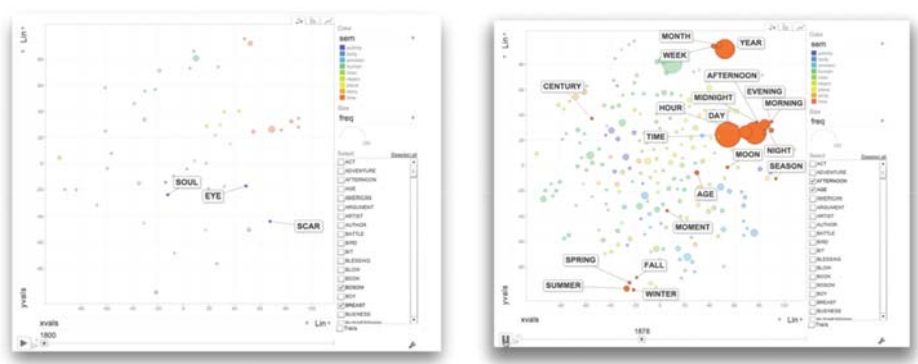

Figure 9. Animated scatterplots (Hilpert et al., 2015)

Martin et al. (2009) introduces a visualization technique for detecting and showing specifically different stages of semantic change based on variability-based neighbor clustering (VNC). VNC is essentially a hierarchical clustering, but it takes into account the temporal ordering of the data. By imposing the rule that merge should be performed only when data points are immediately adjacent, VNC obtains a dendrogram that shows distances among larger clusters. As denoted in Figure (10), the plot shows substantial distances between three large clusters, indicating three stages in the diachronic development.

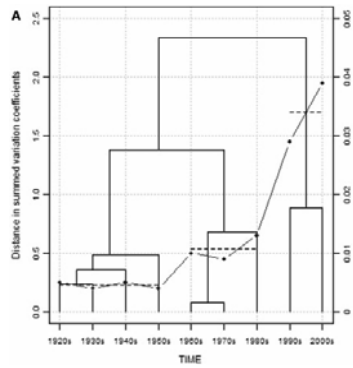

Figure 10. VNC dengrogram (Martin et al., 2009)

## 8 Conclusions

Word meanings are in the state of eternal change. Researchers in both theoretical linguistics and computational linguistics have long been aware of the phenomena, but only in the last decade were the researchers able to study semantic change in large scale corpus with computational approaches. The field is still at its initial phase. The framework presented in the paper summarizes the current research approaches in the field, showing that most studies are focused on testifying hypotheses proposed in theoretical studies of semantic change. The issues discussed in the field, such as the categorization of semantic change, the relationship between frequency and semantic change and the relationship between polysemy and semantic change, have long been discussed in theoretical studies. The models of semantic change, such as the S-shaped curve, are also proposed in theoretical studies. However, semantic change computation possesses advantages that cannot be obtained in the past. It can gather evidence of semantic change in large scale and provide advanced data visualization techniques. From this perspective, the studies of semantic change computation are expected not only to testify existent hypotheses and rules of semantic change, but also to uncover new rules and regularities. Recent studies such as Hamilton et al. (2016) and Dubossarsky et al. (2017) represent such a trend.

This review also shows that there has been no reported application of semantic change computation techniques in NLP tasks, although researchers are aware of the potential of semantic change knowledge in these tasks. In the studies, proposals of NLP application are often briefly mentioned at the introductory parts or the conclusion parts. For instance, Mitra et al. (2015) proposed in their conclusions that novel sense detection can be used to enhance semantic search by increasing the efficiency of word sense disambiguation. In addition, current studies of semantic change computation are more biased towards language in history while NLP applications are more interested in modern and contemporary language use. One piece of evidence for this observation lies in the use of Google Books Ngram corpus. The corpus spans several hundreds of years and is often studied with decades as time interval. However, semantic change can happen within years, or even months. For instance, Tang et al. (2016) show that the metaphorical semantic change of a Chinese word *tou$^{51}$ming$^{35}$* become highly conventionalized within two or three years. To meet the demand of NLP tasks, more research should be directed to short-term semantic changes instead of long-term ones.

The literature has proposed various methods for diachronic word sense characterization and change modelling. Initial comparison has been conducted to show the merits and demerits of these approaches. For instance, Hamilton et al. (2016) show that Skim-Gram Model and general SVD perform differently on different corpus, and PPMI is noticeably worse than the two approaches. But the experiment is conducted on a small evaluation data (less than 40 words). Comprehensive comparison on large scale evaluation data is still needed to validate discoveries in these studies. In addition, there seems to be a tendency to integrate different steps in diachronic word sense characterization into one model, as in the Bayesian model SCAN (Frermann et al., 2016), the Topics-Over-Time model (Wang et al., 2006) and the model based on word embedding (Yao et al., 2017). This tendency also awaits further justification.

Semantic change is one of the most evading topics in the studies of language change, and in linguistic studies in general. But recent advance in the field of semantic change computation shows that it is possible to address the issue with computational approaches and to testify, develop rules and mechanisms of semantic change or even to uncover new rules and mechanisms.